\documentclass[conference]{IEEEtran}
\IEEEoverridecommandlockouts
\usepackage[utf8]{inputenc}
\usepackage[T1]{fontenc}
\usepackage{cite}
\usepackage{amsmath,amssymb,amsfonts}
\usepackage{graphicx}
\usepackage{textcomp}
\usepackage{xcolor}
\usepackage{booktabs}
\usepackage{hyperref}
\usepackage{float}
\usepackage{mathptmx}

\hypersetup{colorlinks=true, linkcolor=blue, urlcolor=blue, citecolor=blue}

\begin{document}

\title{Per-Group Error, Not Total MSE:\\Fine-Tuning Vision-Language-Action Models\\for 11-DoF Mobile Manipulation}

\author{
\IEEEauthorblockN{Pau Montagut Bofi, Mario García Blasco, Tessa Pulli, Markus Vincze}
\IEEEauthorblockA{Vision for Robotics Lab (V4R), TU Wien, Vienna, Austria}
}

\maketitle

\begin{abstract}
Fine-tuning Vision-Language-Action (VLA) models for mobile manipulators with heterogeneous joint spaces can produce a counterintuitive result: the checkpoint with the lowest aggregate MSE is not the one that performs best on the real robot. We argue this is a predictable consequence of collapsing heterogeneous joint groups (arm, gripper, head, wheeled base) into a single metric, where easy-to-predict joints can mask joints that still fail.

We fine-tune SmolVLA (450M, action-expert only) on the 11-DoF Toyota HSR and compare it against $\pi_{0.5}$ (3.3B), a stronger pretrained baseline. Per-group analysis exposes two patterns: in SmolVLA, the mobile base converges slowest and limits overall performance. In expert-only fine-tuning of $\pi_{0.5}$ (training only the action head, backbone frozen), total MSE drops below the baseline but arm accuracy degrades. On 60 real-robot trials (20 per model), $\pi_{0.5}$~80k (4.0/4) significantly outperforms both fine-tuned variants (expert-only 3k: 3.75/4; HSR-SmolVLA: 3.5/4; Mann--Whitney $p \leq 0.010$), despite expert-only 3k having the lowest total MSE. This separation is most consistent with the offline arm-group error, not total MSE or base-group error. We conclude that per-group error is a more reliable signal than total MSE for checkpoint selection on robots with heterogeneous action spaces. Code: \url{https://github.com/paumontagut/per-group-mse-vla}.
\end{abstract}

\section{Introduction}

Deploying a pretrained Vision-Language-Action (VLA) model on a mobile manipulator is where published benchmarks meet their limits. Most VLAs, including $\pi_0$~\cite{pi0}, SmolVLA~\cite{smolvla}, and OpenVLA~\cite{openvla}, are trained on fixed arms with 6 or 7 Cartesian degrees of freedom, whereas platforms such as the Toyota Human Support Robot (HSR) require controlling 11 joints that span arm, gripper, head, and a wheeled base, a combination absent from most pretraining corpora.

The heterogeneity of these joint groups, from the binary gripper to the continuous base motion, matters beyond model architecture: when the 11 dimensions are aggregated into a single MSE, groups that are easy to learn can mask groups that are not. A checkpoint can look competitive on total MSE and still behave poorly on the robot because a deployment-critical group is underfit. As we show, which group that is depends on the model and the training regime.

\begin{figure}[t]
\centering
\includegraphics[width=\columnwidth]{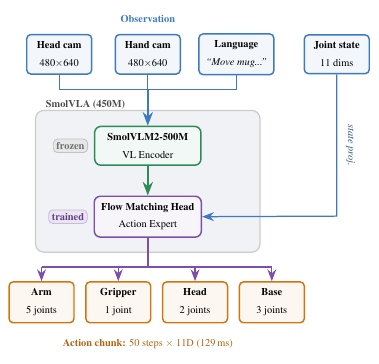}
\caption{End-to-end pipeline: pretraining SmolVLA on HSR teleoperation data, task-specific fine-tuning on a pick-up task (producing HSR-SmolVLA), and offline + real-robot evaluation.}
\label{fig:pipeline}
\end{figure}

\noindent{}We address this through: (1)~a two-phase fine-tuning pipeline for SmolVLA with native 11D output; (2)~a comparative analysis against $\pi_{0.5}$ (3.3B), including expert-only fine-tuning; (3)~an error decomposition by joint group (arm, gripper, head, base) that reveals per-group training dynamics; and (4)~preliminary real-robot validation on the Toyota HSR. Fig.~\ref{fig:pipeline} summarizes the full pipeline.

\section{Related Work}

\textbf{VLA models.} Recent VLA architectures couple a vision-language backbone with an action generation head. OpenVLA~\cite{openvla} (7B) tokenises 7-DoF Cartesian actions and fine-tunes a vision-language model on large robotic corpora. RT-2~\cite{rt2} transfers web-scale multimodal pretraining to action generation. Both are designed for fixed-arm robots and do not generalise naturally to mobile manipulation. The $\pi_{0}$~\cite{pi0} and $\pi_{0.5}$~\cite{pi05} families introduce flow-matching action heads and explicitly include mobile manipulators in their pretraining corpora, making them better suited to our setting. SmolVLA~\cite{smolvla} is a lightweight 450M-parameter alternative with configurable action dimensionality, making it natively compatible with the HSR's 11-DoF joint space without architectural changes. A complementary line of work adapts fixed-arm VLAs to mobile manipulation without retraining (e.g., MoManipVLA~\cite{momanipvla}). We instead fine-tune natively 11-DoF VLAs and study which offline signal best predicts real-robot performance.

\textbf{Non-VLA imitation learning.} ACT~\cite{act} predicts action chunks with a transformer, and Diffusion Policy~\cite{diffpolicy} models action sequences via iterative denoising. Both focus on single fixed-arm tasks and report task success, not per-group error. Their chunked-action output format directly inspired SmolVLA and the $\pi$ family.

\textbf{Loss balancing in multi-output regression.} When a regression model predicts outputs with very different scales, as is the case with our 11-DoF action space, gradients can be dominated by the largest-magnitude components, causing smaller groups to converge more slowly. Uncertainty weighting~\cite{kendall} and GradNorm~\cite{gradnorm} address this automatically. We identify this imbalance as a likely driver of the per-group training dynamics we observe, and leave applying these techniques to future work.

\textbf{Per-group error in VLA evaluation.} To our knowledge, the VLA literature reports evaluation as aggregate MSE or task success, with no systematic decomposition across joint groups. Our work fills this gap for heterogeneous action spaces.

\textbf{The gap for mobile manipulators.} Taken together, these works leave two gaps for mobile manipulators like the HSR: architectures designed for fixed-arm platforms must be adapted, and standard aggregate metrics may not capture the full picture on heterogeneous embodiments. We address both: native 11-DoF output in SmolVLA (architectural), and per-group error decomposition (evaluation).

\section{Method}

\subsection{Models}

\textbf{SmolVLA}~\cite{smolvla} is a 450M-parameter VLA built on SmolVLM2-500M. It generates continuous actions via flow matching with configurable dimensionality (set to 11 for the HSR). Actions are predicted as chunks of 50 future steps. The released SmolVLA checkpoint was pretrained on LeRobot community data, predominantly fixed-arm teleoperation without mobile-base control. We train the action expert via the LeRobot~\cite{lerobot} framework, with the SmolVLM2-500M backbone frozen.

\textbf{$\pi_{0.5}$}~\cite{pi05} is a 3.3B-parameter VLA based on PaliGemma2 (3B vision-language backbone) with a diffusion-based expert action head ($\sim$300M parameters). It was pretrained on diverse manipulation and navigation data spanning multiple embodiments and including mobile base control, a substantially broader corpus than the HSR-only subset we use to pretrain SmolVLA.

\subsection{Training}

\textbf{SmolVLA.} Phase~1: pretraining on a 109,269-episode subset (531 tasks) of the ICRA 2026 workshop dataset~\cite{airoa_icra} for 20k steps, batch size~32. Phase~2: task-specific fine-tuning on a six-task pick-up release from the same workshop (3,971 episodes) for 50k steps, batch size~32, saving every 5k. We refer to the final fine-tuned model as \emph{HSR-SmolVLA}. All training on one RTX~3090 (24\,GB).

\textbf{$\pi_{0.5}$.} We use pretrained baselines at 80k and 100k steps from the workshop dataset~\cite{airoa_icra}. For expert-only fine-tuning, we freeze the PaliGemma2 backbone ($\sim$3B parameters) and train only the expert action head ($\sim$300M parameters) for up to 5k steps. This is the most memory-efficient strategy available: full fine-tuning exceeds 24\,GB VRAM, and LoRA is incompatible with the architecture of the provided checkpoint.

The two model families differ in pretraining corpus: SmolVLA is pretrained only on our 109k-episode HSR-only subset of the workshop dataset, while the $\pi_{0.5}$ baselines were pretrained on a substantially larger, multi-embodiment corpus. The comparison thus contrasts two checkpoints as they are available to practitioners, not two architectures under matched training budgets.

\subsection{Per-Joint-Group Error Analysis}

Standard aggregate MSE obscures important dynamics in heterogeneous action spaces. We decompose the 11D action vector into four functional groups $\mathcal{G} = \{\text{arm}, \text{gripper}, \text{head}, \text{base}\}$ and compute:
\begin{equation}
\text{MSE}_g = \frac{1}{|g|} \sum_{j \in g} (\hat{a}_j - a_j)^2, \quad g \in \mathcal{G}
\end{equation}
where $\hat{a}_j$ is the predicted action and $a_j$ the ground truth for joint~$j$. This reveals which joint groups benefit from fine-tuning and which remain bottlenecks, information invisible in aggregate metrics.

\section{Experiments}

\subsection{Offline Evaluation Protocol}

We evaluate each checkpoint on 10 held-out episodes (20 uniformly sampled frames per episode, 200 frames total). For each frame, we run single-step inference given the head and hand camera images, robot state, and task instruction, and compute per-group MSE between predicted and ground-truth actions.

\subsection{SmolVLA Results}

\begin{figure}[t]
\centering
\includegraphics[width=\columnwidth]{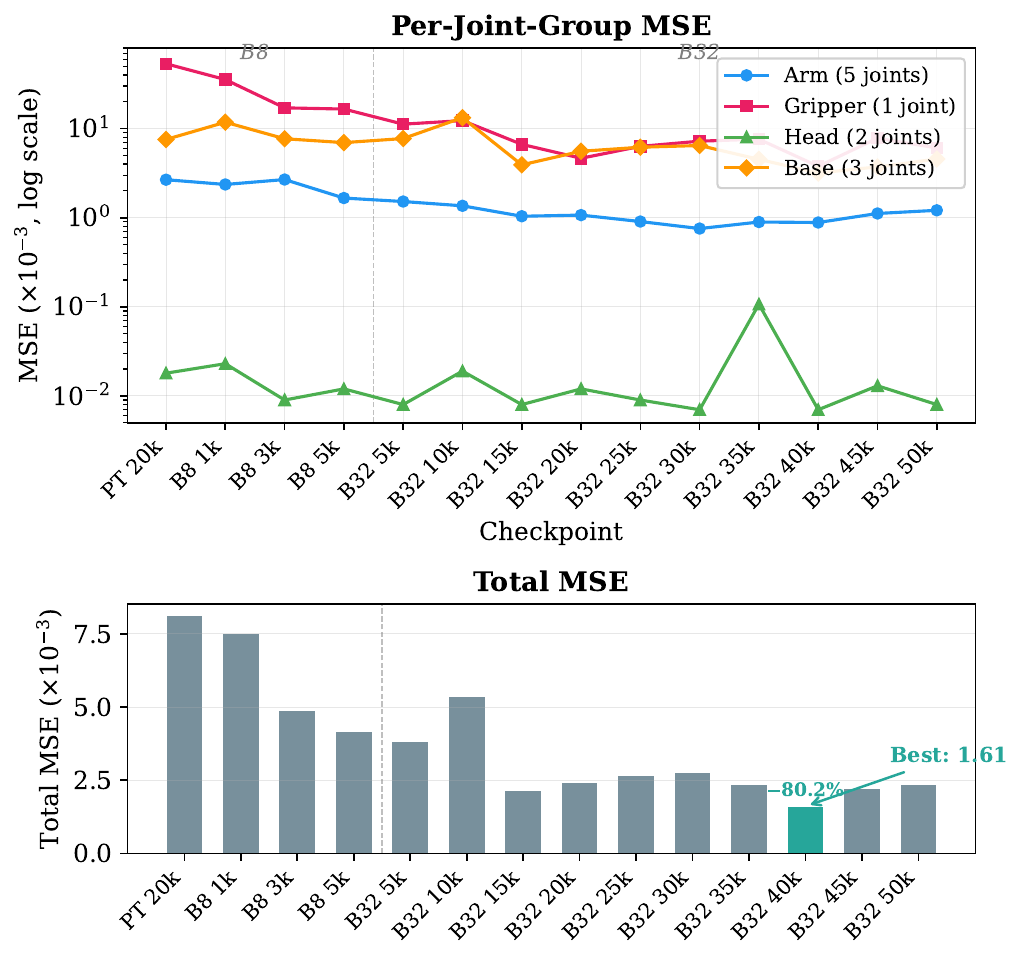}
\caption{Per-joint-group MSE over SmolVLA training. Gripper error falls fastest. Base error is the last to converge and sets the ceiling on total MSE.}
\label{fig:mse_curves}
\end{figure}

Table~\ref{tab:smolvla} reports per-group MSE ($\times 10^{-3}$) for SmolVLA checkpoints along the training trajectory, and Fig.~\ref{fig:mse_curves} visualises the full curve. The pretrained model (end of Phase~1) achieves a total MSE of 8.12. Fine-tuning on the pick-up task reaches its minimum at 40k steps (MSE~1.61, $-80.2\%$). We refer to this checkpoint as HSR-SmolVLA. The trajectory is not strictly monotonic (MSE rises at B32~20k before falling again) and flattens after 40k, with B32~50k slightly worse: the signature of a model that has entered the overfitting regime for the available training data. To keep the table compact we report a representative subset of checkpoints. The full curve in Fig.~\ref{fig:mse_curves} includes every 5k-step checkpoint from 5k to 50k.

\begin{table}[t]
\centering
\caption{SmolVLA per-group MSE ($\times 10^{-3}$). Best in bold.}
\label{tab:smolvla}
\begin{tabular}{@{}lccccc@{}}
\toprule
\textbf{Checkpoint} & \textbf{Arm} & \textbf{Grip.} & \textbf{Head} & \textbf{Base} & \textbf{Total} \\
\midrule
Pretrained 20k       & 2.67 & 53.34 & 0.02 & 7.54 & 8.12 \\
B32 5k               & 1.52 & 11.20 & 0.01 & 7.73 & 3.82 \\
B32 15k              & 1.04 &  6.65 & 0.01 & 3.92 & 2.15 \\
B32 20k              & 1.07 &  4.64 & 0.01 & 5.56 & 2.43 \\
\textbf{HSR-SmolVLA (B32 40k)} & \textbf{0.88} & \textbf{3.77} & \textbf{0.01} & \textbf{3.18} & \textbf{1.61} \\
B32 50k              & 1.21 &  6.06 & 0.01 & 4.55 & 2.34 \\
\bottomrule
\end{tabular}
\end{table}

The decomposition reveals distinct convergence rates: the gripper improves 92.9\% (53.34$\to$3.77), while the base improves only 57.8\% (7.54$\to$3.18) and remains the dominant error source. Head MSE is already very low after pretraining. However, we caution that this likely reflects the nature of the dataset (in the pick-up task, the target objects are always within the initial field of view, so the head rarely needs to reorient) rather than the problem being solved in a general sense.

\subsection{$\pi_{0.5}$ Baselines and Expert-Only Fine-Tuning}

Table~\ref{tab:pi05} compares the $\pi_{0.5}$ baselines with our expert-only fine-tuning. The 80k baseline achieves MSE~1.04, outperforming SmolVLA's best (1.61) by 35.4\%. Notably, continuing training to 100k degrades performance (1.21, $+16\%$), indicating overfitting even in the pretrained model.

Expert-only fine-tuning from the 80k checkpoint improves total MSE to 0.95 at 3k steps ($-9.0\%$ vs.\ baseline), primarily through a large gripper improvement (4.31$\to$2.21). However, arm MSE degrades (0.30$\to$0.59), suggesting that the optimization is dominated by the joint groups with larger error magnitudes (gripper and base): reducing them trades off accuracy on the already well-predicted arm. At 5k steps, gripper MSE spikes to 4.21, erasing gains: rapid overfitting consistent with the small fine-tuning dataset.

\begin{table}[!h]
\centering
\caption{$\pi_{0.5}$ per-group MSE ($\times 10^{-3}$). Best in bold.}
\label{tab:pi05}
\begin{tabular}{@{}lccccc@{}}
\toprule
\textbf{Checkpoint} & \textbf{Arm} & \textbf{Grip.} & \textbf{Head} & \textbf{Base} & \textbf{Total} \\
\midrule
Baseline 80k           & 0.30 & 4.31 & 0.05 & 1.85 & 1.04 \\
Baseline 100k          & 0.33 & 5.46 & 0.22 & 1.91 & 1.21 \\
\midrule
Expert-only 1k         & 0.48 & 2.59 & 0.06 & 1.87 & 0.97 \\
\textbf{Expert-only 3k}& 0.59 & \textbf{2.21} & 0.06 & \textbf{1.72} & \textbf{0.95} \\
Expert-only 4k         & 0.54 & 2.20 & 0.09 & 1.79 & 0.95 \\
Expert-only 5k         & 0.52 & 4.21 & 0.07 & 1.89 & 1.15 \\
\bottomrule
\end{tabular}
\end{table}

\subsection{Real-Robot Validation}

\begin{figure}[t]
\centering
\includegraphics[width=\columnwidth]{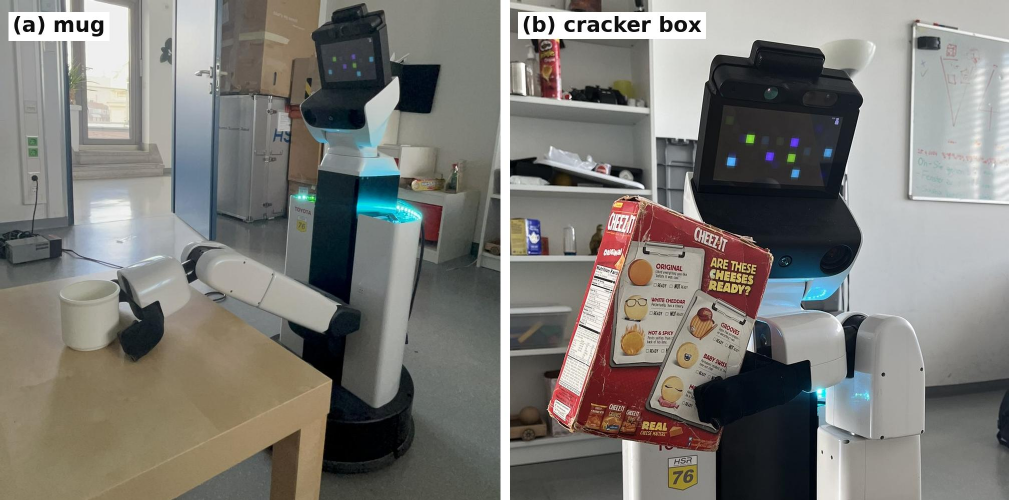}
\caption{Real-robot evaluation tasks on the Toyota HSR: (a)~approaching a ceramic mug, (b)~grasping a Cheez-It cracker box. Each model is run for 20 trials across the two objects.}
\label{fig:tasks}
\end{figure}

We conducted 60 trials on the Toyota HSR (20 per model) on a pick-up task (mug and cracker box; Fig.~\ref{fig:tasks}). Each trial was graded on a 4-point rubric defined in Table~\ref{tab:robot}: from~1 (the robot does not approach the target) to~4 (full task completion).

\begin{table}[t]
\centering
\caption{Real-robot trials (20 per model) graded on a 4-point rubric. Score key: 4 = complete task; 3 = touches the object but does not lift; 2 = approaches but does not grasp; 1 = does not approach. Columns $n_k$ count trials receiving score $k$.}
\label{tab:robot}
\begin{tabular}{@{}lcccccc@{}}
\toprule
\textbf{Model} & \textbf{MSE} & $n_1$ & $n_2$ & $n_3$ & $n_4$ & \textbf{Mean} \\
\midrule
$\pi_{0.5}$ 80k             & 1.04 & 0 & 0 & 0  & 20 & \textbf{4.0/4} \\
Expert-only 3k              & 0.95 & 0 & 0 & 5  & 15 & 3.75/4 \\
HSR-SmolVLA (40k)           & 1.61 & 0 & 2 & 6  & 12 & 3.5/4 \\
\bottomrule
\end{tabular}
\end{table}

$\pi_{0.5}$~80k achieves a perfect 20/20 on the rubric, while the two fine-tuned models perform worse: expert-only 3k at 3.75/4 (15/20 lifts) and HSR-SmolVLA at 3.5/4 (12/20). A Mann--Whitney test confirms that $\pi_{0.5}$~80k significantly outperforms both expert-only 3k ($p = 0.010$) and HSR-SmolVLA ($p = 0.001$). Notably, expert-only 3k underperforms the untouched baseline despite having lower total MSE (0.95 vs.\ 1.04) and lower base-group error (1.72 vs.\ 1.85). This disconnect lines up with the offline arm-group error (0.59 vs.\ 0.30), which is the only offline metric consistent with the deployment ranking. The 0.25-point gap between expert-only 3k and HSR-SmolVLA is not statistically significant ($p = 0.128$), so these two should not be read as cleanly ordered. Qualitatively, most HSR-SmolVLA failures were accurate in navigation but miscalibrated in grasping: in the two trials scored~2, the robot approached the table confidently but attempted the grasp approximately 20\,cm away from the target object, closing the gripper in empty space and lifting nothing. This pattern is consistent with SmolVLA's relatively high residual arm error ($0.88\times 10^{-3}$) compared with $\pi_{0.5}$ ($0.30\times 10^{-3}$).

\section{Discussion and Conclusion}
\label{sec:discussion}

The most concrete lesson from our experiments is that \emph{total MSE is not, by itself, a reliable criterion for selecting a checkpoint destined for a real robot}, and that the joint group that dominates the residual error varies across models and training regimes. Within SmolVLA, total MSE falls 80\% during fine-tuning, but Fig.~\ref{fig:mse_curves} shows that this average hides a gripper that converges ten times faster than the mobile base. The base is the last group to converge and sets the ceiling on the aggregate. This is the sense in which the \emph{mobile base is the bottleneck of SmolVLA fine-tuning}. The underlying SmolVLA checkpoint was pretrained on LeRobot community data, dominated by fixed-arm teleoperation and with no mobile-base exposure. Our Phase-1 training on 109k HSR episodes then adds the base modality, but 20k steps on this HSR-only subset followed by 50k fine-tuning steps on 3,971 episodes are not enough to bring the base-group error down to the level of the gripper or head. The picture is different for $\pi_{0.5}$, which was pretrained on a substantially larger and more diverse corpus including mobile base control across multiple embodiments: its residual is dominated not by the base but by the gripper (4.31 vs.\ 1.85 per-group MSE at 80k). On 11-DoF mobile platforms, base-group convergence appears sensitive to the \emph{scale and diversity} of pretraining, not merely to its presence.

The expert-only fine-tuning of $\pi_{0.5}$ exposes a second, more subtle case of this disconnect and, in doing so, identifies which group matters for deployment. Expert-only 3k attains the lowest total MSE in the paper (0.95 vs.\ 1.04) and even improves on the base-group error (1.72 vs.\ 1.85), yet it loses to the untouched baseline on the robot (3.75/4 vs.\ 4.0/4, $p = 0.010$). The per-group view makes this intelligible: the gains concentrate on the gripper and the base, both of which had large residual error, while the arm (already well-predicted at 0.30) degrades to 0.59. In five of the twenty trials the robot correctly approached the object and closed the gripper around it, but failed to lift cleanly, consistent with an arm policy that has lost calibration. The significant baseline-vs-fine-tuned separation on the robot lines up with the offline arm-group error, not with the base-group error or the total. This recasts the bottleneck picture: the \emph{training-time} bottleneck for SmolVLA is the base group (slowest to converge), while the \emph{deployment-time} bottleneck in our expert-only experiment is the arm group (degraded under fine-tuning). The shared mechanism is a familiar one in multi-output regression: when per-group loss magnitudes are imbalanced by one to two orders of magnitude, an unweighted objective trades capacity away from small-magnitude groups. Uncertainty weighting~\cite{kendall} or gradient balancing~\cite{gradnorm} would let the procedure protect the arm while still improving the gripper and base. We leave an empirical study of this to future work.

For other teams deploying VLAs on the HSR or similarly heterogeneous embodiments, three guidelines follow from our observations. First, report per-group error alongside total MSE: a checkpoint can look competitive on the aggregate and still be a poor match for deployment. Second, identify the dominant group in each regime before trusting it as a deployment proxy: in our SmolVLA results the bottleneck is the base, while in the $\pi_{0.5}$ expert-only comparison the arm turns out to be the discriminative group. Third, keep fine-tuning short when task data is scarce: both SmolVLA (beyond 40k steps) and $\pi_{0.5}$ expert-only (beyond 3k steps) enter an overfitting regime visible in the total, and per-group error additionally reveals which group degrades first.

\textbf{Limitations.} Our real-robot evaluation covers 60 trials (20 per model) under controlled initial conditions on a pick-up task, sufficient for statistical comparison of the baseline-vs-fine-tuned contrast (Mann--Whitney $p \leq 0.010$), but not to resolve the gap between the two fine-tuned models ($p = 0.128$). Generalization to other task types and embodiments remains future work. Offline MSE is a frame-level metric and does not capture errors that compound over trajectories. Finally, the comparison between SmolVLA and $\pi_{0.5}$ involves different pretraining corpora and is therefore not a controlled study of model architectures. It reflects the practical choice between a checkpoint a small team can produce and a larger pretrained one available as a workshop baseline.

\textbf{Conclusion.} Per-group error, not aggregate MSE, is the more faithful signal for selecting VLA checkpoints destined for deployment on heterogeneous action spaces. Which joint group dominates is model- and regime-dependent: the mobile base during SmolVLA fine-tuning, the arm under $\pi_{0.5}$ expert-only fine-tuning. In both regimes, it is the per-group view that tracks real-robot performance.


\end{document}